\def\year{2020}\relax
\documentclass[letterpaper]{article} 
\usepackage{aaai20}  
\usepackage{times}  
\usepackage{helvet} 
\usepackage{courier}  
\usepackage[hyphens]{url}  
\usepackage{graphicx} 
\usepackage{amsmath, amssymb}
\usepackage[inline]{enumitem}
\usepackage[noend]{algpseudocode}
\usepackage{algorithmicx,algorithm}
\usepackage{color}
\usepackage{transparent}
\usepackage{relsize}
\newcommand{\citep}[1]{\cite{#1}}
\setlist[description]{labelindent=\parindent,leftmargin=\parindent}
\usepackage{graphicx}  
\usepackage{xcolor}
\usepackage{mathtools}
\newcommand{\defeq}{\vcentcolon=}

\def\state{s}

\def\obs{o}

\def\ObsSp{\mathcal{O}}

\def\act{a}

\def\StateSp{\mathcal{S}}
\def\ActSp{\mathcal{A}}

\def\Trans{T}

\def\Rew{R}

\def\R{\mathbb{R}}

\newcommand{\policy}{\pi}

\newcommand{\goal}{g}

\newcommand{\Loss}{\mathcal{L}}

\newcommand{\encoder}{\psi}
\newcommand{\cmd}{c}
\newcommand{\Cmd}{\mathcal{C}}
\newcommand{\hact}{\hat{\act}}
\newcommand{\hactst}{\hat{\act}_{t,s}}
\newcommand{\hactbt}{\hat{\act}_{t,b}}
\newcommand{\actst}{\act_{t,s}}
\newcommand{\actbt}{\act_{t,b}}
\newcommand{\ActSpS}{\ActSp_s}
\newcommand{\ActSpB}{\ActSp_b}
\newcommand{\toppolicy}{\hat{\phi}}
\newcommand{\goalencoder}{\phi}
\newcommand{\hgoal}{\hat{\goal}}
\newcommand{\lclose}{y}
\title{Learning from Interventions using Hierarchical Policies for Safe Learning 
}
\author{
  Jing Bi\textsuperscript{\rm 1},
  Vikas Dhiman\textsuperscript{\rm 2},
  Tianyou Xiao\textsuperscript{\rm 1},
  Chenliang Xu\textsuperscript{\rm 1} \\
  \textsuperscript{\rm 1}University of Rochester\\
  \textsuperscript{\rm 2}University of California San Diego\\
  jbi5@ur.rochester.edu, vdhiman@ucsd.edu\\
  txiao3@u.rochester.edu, chenliang.xu@rochester.edu
  }
\makeatletter
\def\thetitle{\@title}
\def\theauthor{Jing Bi, Vikas Dhiman, Tianyou Xiao, Chenliang Xu}
\makeatother
\begin{document}
\maketitle
\begin{abstract}
Learning from Demonstrations (LfD) via Behavior Cloning (BC) works well on multiple complex tasks. However, a limitation of the typical LfD approach is that it requires expert demonstrations for all scenarios, including those in which the algorithm is already well-trained. The recently proposed Learning from Interventions (LfI) overcomes this limitation by using an expert overseer. The expert overseer only intervenes when it suspects that an unsafe action is about to be taken. Although LfI significantly improves over LfD, the state-of-the-art LfI fails to account for delay caused by the expert's reaction time and only learns short-term behavior. We address these limitations by 1) interpolating the expert's interventions back in time, and 2) by splitting the policy into two hierarchical levels, one that generates sub-goals for the future and another that generates actions to reach those desired sub-goals. This sub-goal prediction forces the algorithm to learn long-term behavior while also being robust to the expert's reaction time. Our experiments show that LfI using sub-goals in a hierarchical policy framework trains faster and achieves better asymptotic performance than typical LfD.
\end{abstract}
\section{Introduction}
Methodologies for learning control policies can be categorized into Reinforcement Learning (RL) and Imitation Learning (IL).
RL depends on the availability of a well-designed reward function, while IL depends on demonstrations from an expert.
Recently, RL has had success in simulated environments and games~\citep{MnKaSiNATURE2015,SiHuMaNATURE2016}.
However, in many real-world robotic applications like autonomous driving, it is challenging to design explicit reward functions. Moreover, training by trial-and-error in the real world is highly unsafe. 
For these reasons, IL is often preferable to RL.
\begin{figure}[t]%
  \includegraphics[width=\columnwidth]{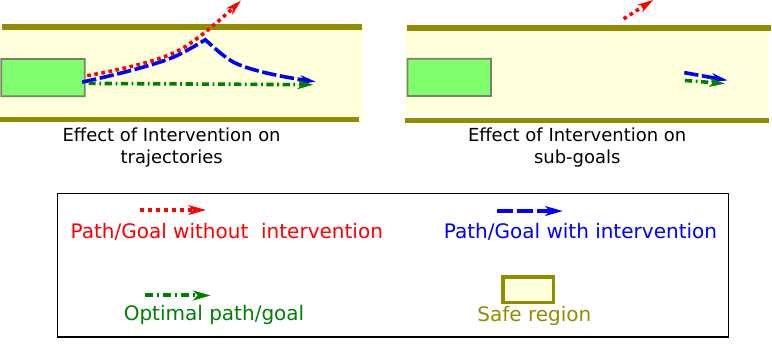}%
  \caption{
  An autonomous driving example to show the advantage of
    intervening on sub-goals rather than trajectories.
    An expert-overseer intervenes whenever the car is about to go outside the
    safe region, with some delay.
    The figure on the left shows three trajectories: without intervention, with
    intervention, and the optimal trajectory.
    There is a difference between the intervened trajectory and the desired trajectory due to the expert's reaction time.
    However, by formulating the intervention in terms of sub-goals, we can minimize this discrepancy, as shown in the right figure.
    }%
  \label{fig:paper-summary}%
\end{figure}%

There are two main categories of Imitation Learning: Inverse Reinforcement Learning (IRL) and Behavior Cloning (BC), which is also known as Learning from Demonstration (LfD). In IRL, the algorithm learns a reward function from expert demonstrations, then learns the optimal policy through trial and error.  On the other hand, BC directly learns the policy to match expert demonstration, without ever learning the reward function. In safety-critical scenarios, BC is preferred over IRL because it requires no trial-and-error during training. However, BC requires a large number of data samples, as it depends on the demonstration data in all scenarios, even if the algorithm has learned the scenarios satisfactorily. 

\cite{saunders_trial_2018} and \cite{waytowich_cycle--learning_2018} have recently introduced the framework of Learning from Interventions (LfI) that allows for safe RL in the presence of a human overseer. This approach guarantees the safety and is shown to need fewer samples from an expert than LfD. State-of-the-art approaches in both LfD and LfI learn reactive behaviors~\citep{tai2016survey} and very little long-term behavior.  Moreover, no LfI framework so far considers the reaction delay in modeling expert supervision. We find that the expert only signals after a non-negligible amount of delay and that compensating for these delays leads to improvements in both sample efficiency and performance.
\begin{figure}[t]
    \centering
    \includegraphics[width=\columnwidth]{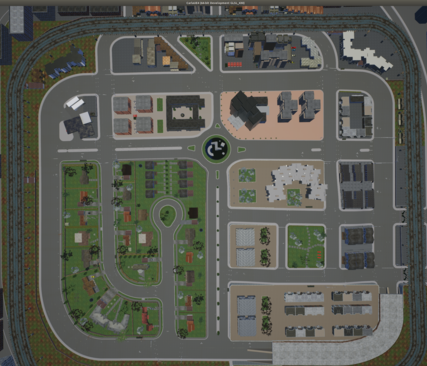}
    \caption{Top view of the map in CARLA~\cite{CARLA} simulator where experiments were conducted.
    }
    \label{fig:Carla}
\end{figure}
\begin{figure*}[t]%
  \def\imgfrac{0.142}
  \centering
  \includegraphics[width= 2.1\columnwidth]{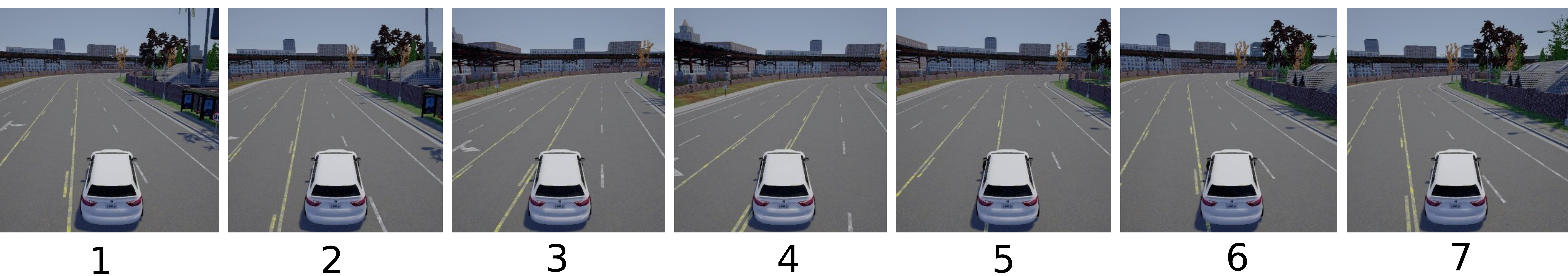}%
  \caption{ Snapshots from the CARLA simulator~\cite{CARLA}  showing that the car is already out of lane
  by image (1), but that the expert does not intervene until image (4)  due to reaction delay and veers it back to desired trajectory (4-7). }%
  \label{fig:carla-sim-interventions}%
\end{figure*}%

To address both of these challenges in LfI, we employ ideas from Hierarchical Reinforcement Learning~\citep{kulkarni2016hierarchical,savinov2018semi} by modeling our policy as multiple hierarchical policies operating at different time-scales, one focused on the long-term and another focused on the short-term.
Also, as demonstrated in Fig.~\ref{fig:paper-summary}, long term predictions are less likely to suffer from an expert's reaction delay. Therefore our model can compensate for the expert's reaction time.

Our proposed approach is as follows.
We start with conditional IL \citep{codevilla_end--end_2018} as our
non-hierarchical baseline. All our policies are \textit{conditioned} on a
command signal, which represents a high-level instruction, such as whether to turn left or right at an intersection. 
We split the policy into two hierarchy levels, where the top-level policy
predicts a \textit{sub-goal} to be achieved $k$ steps ahead in the future.
The bottom-level policy chooses the best action to achieve the \textit{sub-goal}
generated by the top-level policy.
We assume any achieved state is a desirable goal from the perspective of a past state - embeddings from the present safe state are used as a sub-goal for the past-state, similar to in Hindsight Experience Replay~\citep{andrychowicz2017hindsight}.
Since we work with images as our state space, we do not generate sub-goals in
the image space, as it is unnecessary to predict future images at pixel-level
detail when we only want to reach the sub-goal. Instead, we
generate intermediate sub-goal embeddings using Triplet Networks~\citep{hoffer_deep_2014}.
The details of our approach are discussed further in the Methods section.

We evaluate our proposed system and relevant baselines on the task of
autonomous driving in a realistic simulated environment of urban driving. The
agent is spawned in a random location on the map. It navigates autonomously
according to a controller's high-level commands. To be successful at this task, the agent must
stay in its lane and not crash into buildings or other vehicles. Our results
suggest that by combining expert intervention and hierarchical policies, our method helps the agent in two fundamental ways:
Firstly, it helps the agent to improve its policy online without taking unsafe actions.
Secondly, hierarchical policies force the agent to predict and plan longer-term behaviors, leading to more effective use of data. 
  
To summarize, our contributions are three-fold:
\begin{enumerate}
 \item We propose a \textit{new problem formulation} of
   Learning from Interventions that incorporates the expert's reaction delay.
 \item We propose a \textit{novel algorithm} to address the proposed problem,
   which combines Learning from Interventions with Hierarchical Reinforcement
   Learning.
 \item We propose an interpolation trick, called \textit{Backtracking}, that allows us to use state-action pairs before and after the intervention.
 \item We propose a \textit{novel Triplet-Network based architecture} to train
the hierarchical policies in the absence of ground-truth sub-goals.
\end{enumerate}
We further demonstrate the practical effectiveness of our approach on a real-world RC robotic truck.
Experiments with the physical system demonstrate that our approach can be successfully deployed in the physical world. 

\section{Related Work}

\subsection{Imitation Learning}

The approaches for Imitation Learning can be categorized into Behavior Cloning~\citep{ross_learning_2013,7358076} and Inverse Reinforcement Learning~\citep{abbeel_apprenticeship_2004}. For the scenarios, when high-level command instructions are available, Codevilla~et~al.~\citep{codevilla_end--end_2018} proposed conditional IL method that learns conditional policies based on the high-level command instructions.

Behavior Cloning might lead to unsafe actions, especially in the less frequently sampled states~\citep{survey}. Therefore, in real-world scenarios, the expert's oversight and interaction are necessary to help the agent train without failure and unsafe actions.

The approaches to Learning from Interventions (LfI) can be categorized according to the frequency of the expert's engagement. The expert's engagement varies from high frequency in Learning from Demonstrations~\citep{akgun_keyframe-based_2012,36},  medium frequency in learning from Interventions~\citep{akgun_trajectories_2012,bi_navigation_2018} to low frequency in Learning from Evaluations~\citep{knox_interactively_2009,macglashan_interactive_nodate}.  Recent work has applied LfI to safe RL~\citep{saunders_trial_2018} and IL~\citep{NIPS2016_6391}, which allows the agent to perform a certain task without entering unsafe states. In ~\citep{hilleli2018toward}, a classifier is trained from the expert demonstration that used to separate safe states from undesired states. When an undesired state is detected, another policy is activated to take over actions from the agent when necessary.  In several works~\citep{hilleli2018toward,peng_deepmimic}, human intervention is often used to reshape the reward of RL, which encourages the agent to avoid certain states. The work closest to our approach is~\citep{goecks_efficiently_nodate}, which proposes \textit{Cycle-of-Learning} that combines different interaction modalities into a single learning framework. This method allows the expert to intervene by providing corrective actions when the agent encounters unwanted states and further aggregates these states and actions into the training dataset. The intervention data from failure states help the agent better generalize and overcome the deficiencies in its policy. We further discuss the effect of states not only used to recover from the unwanted states but also the states, which will lead to failure. 

\subsection{Inverse Dynamics}
Learning a supervised policy is known to have ``myopic'' strategies, since it ignores the temporal dependence between consecutive states. In contrast, for making informative decisions, classical planning approaches in robotics can reason far into the future, but often require designing sophisticated cost functions. Savinov~et~al.~\cite{savinov2018semi} propose to directly model Inverse dynamics to represent a trade-off between learning robust policies for long term planning. Formally, the inverse dynamics is probability of taking an action $\act_t$ given the current state $\state_t$ and next states $\state_{t+k}$, i.e. $\mathcal{P}(\act_t|\state_t,\state_{t+k})$. For instance, if $k=1$, the network learns action to perform in one-step transitions with neighbor states. For $k>1$, training data become noisier but sufficient for short-term predicting~\citep{savinov2018semi}. 

Recent works use inverse dynamics to learn features for recognition tasks~\citep{agrawal2015learning} and control tasks~\citep{dosovitskiy_learning_2016}. In ~\citep{agrawal2016learning}, the authors constructed a joint inverse-forward model to learn feature representation for the task of pushing objects. In ~\citep{pathak_curiosity-driven_2017}, an Intrinsic Curiosity Module was introduced to tackle the exploration problem in RL, which uses Inverse Dynamic Model to predict action based on current and future state and a forward model to predict the future state. In this way, the agent explores the unfamiliar states more than familiar ones. The work shows that predicting the future state helps the agent to plan better. We use the inverse dynamics module as the bottom-level policy in our proposed method.

In \citep{hilrl} and \citep{hilrl2}, the high-level algorithm focuses on decomposing the multi-stage human demonstration into primitives actions, and the low-level policy executes a series of actions to accomplished the primitive. In our model, we generate sub-goals that are feature embeddings directly generated by the network, instead of the primitive actions which have precise semantic meanings like ``go to the elevator,'' then ``take the elevator down.'' Therefore, our approach does not need the design of a library of primitive actions.

\section{Problem Formulation}

We consider a set-up similar to~\citep{saunders_trial_2018}: an agent interacting with an environment in discrete time steps.
We assume that this process is a fully observable Goal-conditioned Markov Decision Process $(\StateSp, \ActSp, \Trans, \Rew, \Cmd)$, where $\StateSp$ is the state space, $\ActSp$ is the action space, $\Trans: \StateSp \times \ActSp \mapsto \StateSp$ is the unknown transition model and $\Rew: \StateSp \times \ActSp \times \Cmd \mapsto \R^+$ is the unknown reward function that depends upon the command vector $\cmd_t \in \Cmd$ for specifying the expert's intention.
At each time step $t$, the agent gets an observation $\obs_t \in \ObsSp$ (sequence of images), and a command $\cmd_t$ from the expert, then takes action $\act_t$.
We assume observations provide full information of the current state $\state_t
\in \StateSp$ through a feature extraction mapping $\state_t =
\encoder (\obs_t; \theta_e)$. 
Instead of reward function, we only have access to the expert with
policy $\policy^*: \ObsSp \times \Cmd \mapsto \ActSp$, who can intervene
whenever s/he perceives a probability of catastrophic damage is higher than a
threshold: $p_c^*(\state_{t-k^*}, \act_{t-k^*}) > 1-\delta_{c}$.
So the effective policy followed with expert intervention is given by:
\begin{align}
  \policy_{\text{eff}}(\state_t, \cmd_t) &=
  \begin{cases}
    \policy^*(\state_t, \cmd_t), & \text{ if } p_{c}^*(\state_{t-k^*}, \act_{t-k^*}) > 1-\delta_{c},
    \\
    \policy(\state_t, \cmd_t; \theta), & \text{ otherwise },
  \end{cases}
\end{align}%
where $\policy(\state_t, \cmd_t; \theta)$ is the learned policy so far, and $k^*$ is the unknown expert's reaction delay.
Our aim is to improve the learned $\policy(\state_t, \cmd_t; \theta)$ without any
catastrophic failure:
\begin{align}
  \theta^* = \arg~\min_{\theta} \sum_{t \in \tau_{\text{intervention}}} \|
  \policy_{\text{des}}(\state_t, \cmd_t)
  - \policy(\state_t, \cmd_t; \theta)\|_\ActSp,
\end{align}%
where $\tau_{\text{intervention}}$ is the set of data-samples affected by
interventions and $\policy_{\text{des}}$ is the desired policy which is a
heuristic approximation of $\policy_{\text{eff}}$ that accounts for the expert's reaction delay.

\section{Methods}

\begin{figure}[t]
\def\svgwidth{1.1\linewidth}
   \hspace{0.2in}%
\begingroup%
  \makeatletter%
  \providecommand\rotatebox[2]{#2}%
  \ifx\svgwidth\undefined%
    \setlength{\unitlength}{253.95420825bp}%
    \ifx\svgscale\undefined%
      \relax%
    \else%
      \setlength{\unitlength}{\unitlength * \real{\svgscale}}%
    \fi%
  \else%
    \setlength{\unitlength}{\svgwidth}%
  \fi%
  \global\let\svgwidth\undefined%
  \global\let\svgscale\undefined%
  \makeatother%
  \begin{picture}(1,0.47017086)%
    \put(0,0){\includegraphics[width=\unitlength,page=1]{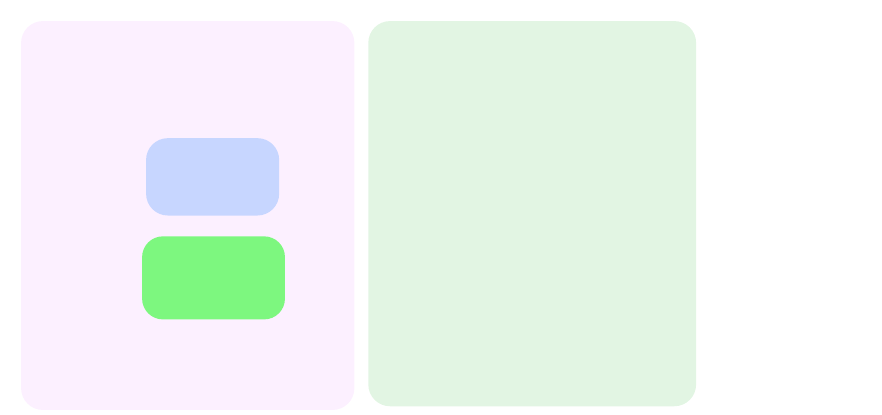}}%
    \put(0.32446235,0.26009255){\color[rgb]{0,0,0}\makebox(0,0)[lt]{\begin{minipage}{0.56703129\unitlength}\raggedright $\state_t$\end{minipage}}}%
    \put(0.04609432,0.15330664){\color[rgb]{0,0,0}\makebox(0,0)[lt]{\begin{minipage}{0.36001984\unitlength}\raggedright $\obs_{t+k}$\end{minipage}}}%
    \put(0.19479441,0.17543781){\color[rgb]{0,0,0}\makebox(0,0)[lt]{\begin{minipage}{0.83188979\unitlength}\raggedright $\goalencoder(\encoder(.))$\\ \end{minipage}}}%
    \put(0.03863807,0.39594998){\color[rgb]{0,0,0}\makebox(0,0)[lt]{\begin{minipage}{0.56703129\unitlength}\raggedright $\cmd_t$\end{minipage}}}%
    \put(0,0){\includegraphics[width=\unitlength,page=2]{main-overall.pdf}}%
    \put(0.49541563,0.38780156){\color[rgb]{0,0,0}\makebox(0,0)[lt]{\begin{minipage}{0.63003477\unitlength}\raggedright $\policy$\end{minipage}}}%
    \put(0.49143794,0.18334803){\color[rgb]{0,0,0}\makebox(0,0)[lt]{\begin{minipage}{0.63003477\unitlength}\raggedright $\policy$\end{minipage}}}%
    \put(0.33859509,0.15316271){\color[rgb]{0,0,0}\makebox(0,0)[lt]{\begin{minipage}{0.63003477\unitlength}\raggedright $\goal_{t+k}$\\ \end{minipage}}}%
    \put(0.36068401,0.43688888){\color[rgb]{0,0,0}\makebox(0,0)[lt]{\begin{minipage}{0.63003477\unitlength}\raggedright $\hgoal_{t+k}$\\ \end{minipage}}}%
    \put(0,0){\includegraphics[width=\unitlength,page=3]{main-overall.pdf}}%
    \put(0.67272147,0.40296197){\color[rgb]{0,0,0}\makebox(0,0)[lt]{\begin{minipage}{0.63003477\unitlength}\raggedright $\hact_{t}$\\ \end{minipage}}}%
    \put(0.66237942,0.20487275){\color[rgb]{0,0,0}\makebox(0,0)[lt]{\begin{minipage}{0.63003477\unitlength}\raggedright $\hact'_{t}$\\ \end{minipage}}}%
    \put(0.77698462,0.2817784){\color[rgb]{0,0,0}\makebox(0,0)[lt]{\begin{minipage}{0.63003477\unitlength}\raggedright $\act_{t}$\\ \end{minipage}}}%
    \put(0.62195652,0.30674055){\color[rgb]{0,0,0}\makebox(0,0)[lt]{\begin{minipage}{0.43588122\unitlength}\raggedright $L_{\text{Bottom}}$\end{minipage}}}%
    \put(0,0){\includegraphics[width=\unitlength,page=4]{main-overall.pdf}}%
    \put(0.21611747,0.2907458){\color[rgb]{0,0,0}\makebox(0,0)[lt]{\begin{minipage}{0.63003477\unitlength}\raggedright $\encoder$\end{minipage}}}%
    \put(0.25754946,0.40842161){\color[rgb]{0,0,0}\makebox(0,0)[lt]{\begin{minipage}{0.63003477\unitlength}\raggedright $\toppolicy_k$\end{minipage}}}%
    \put(0.43286821,0.07524085){\color[rgb]{0,0,0}\makebox(0,0)[lt]{\begin{minipage}{0.35442498\unitlength}\raggedright Bottom-level policy\end{minipage}}}%
    \put(0.0441422,0.07676229){\color[rgb]{0,0,0}\makebox(0,0)[lt]{\begin{minipage}{0.31480613\unitlength}\raggedright Top-level policy\end{minipage}}}%
    \put(0.04609432,0.26671291){\color[rgb]{0,0,0}\makebox(0,0)[lt]{\begin{minipage}{0.36001984\unitlength}\raggedright $\obs_{t}$\end{minipage}}}%
  \end{picture}%
\endgroup%
  \caption{Network diagram for our hierarchical policy. We split our policy into
    two levels of a hierarchy. The top-level policy, $\toppolicy_k$, generates
    sub-goals $\hgoal_{t+k}$ that the bottom-level policy $\policy$ follows.
    We also generate ``true sub-goals'' $\goal_{t+k}$ using a goal encoder
    $\goalencoder$ on future observation $\obs_{t+k}$.
    The loss $\Loss_{\text{Bottom}}$ compares predicted actions $\hact_t$ and
    $\hact'_t$ with true intervention actions $\act_t$.
    We train the goal-encoder and top-policy togther using the TripletNet by
comparing ``true sub-goals'' with randomly generated sub-goals.
  }
  \label{fig:main}
\end{figure}

We present two approaches to address the problem of delayed expert, \emph{hierarchical policies} and \emph{backtracking}.
Please note that both these approaches are not mutually exclusive and can be combined together.

\subsection{Learning from Intervention using Hierarchical policies}
Instead of learning a direct policy network $\policy (\state_t, \cmd_t; \theta)$, we
divide the policy learning into two levels of a hierarchy. 
The top-level policy $\toppolicy_{k}$ learns to predict a sub-goal vector $\hgoal_{t+k}$, $k$ steps ahead, 
while the bottom-level policy $\policy$ learns to generate action to achieve that sub-goal. 
The full policy $\policy (\state_t, \cmd_t;\theta)$ can be written as nested policies:
\begin{align}
  \policy(\obs_t, \cmd_t;\theta) &= \policy (
  \state_t,
  \hgoal_{t+k};
  \theta_b),
  \\
  \hgoal_{t+k} &\defeq \toppolicy_{k}(\state_t, \cmd_t; \theta_t),
  \label{eq:hierarchical-policy}
\end{align}%
where $\state_t = \encoder (\obs_t; \theta_e)$, and $\hgoal_{t+k}$ is the predicted sub-goal by the top-level policy.

We do not have access to the ground truth value of $\hgoal_{t+k}$ from the expert policy $\policy^*$.
In order to learn $\toppolicy_{k}(.;\theta_t)$, we need another network
$\goal_{t+k} \defeq \goalencoder (\state_{t+k}; \theta_g)$, which outputs the
desired sub-goal embedding $\goal_{t+k}$ from an achieved state
$\state_{t+k} = \encoder(\obs_{t+k}; \theta_e)$. 
This strategy is also called Hindsight Experience Replay, because we consider an achieved goal as a desired goal for past observations~\citep{andrychowicz2017hindsight}. 
We train $\toppolicy_{k}$ and $\goalencoder$ together by adapting a Triplet
Network~\citep{hoffer_deep_2014}, which provides a loss that brings closer
the embeddings $\hgoal_{t+k}$ and $\goal_{t+k}$ while pushing away the embedding
$\hgoal_{t+k}$ from a randomly chosen goal embedding $\goal_r =
\goalencoder (\state_r; \theta_g)$. The split of the overall policy into two levels while using the future state as the sub-goal embedding is summarized in Figure~\ref{fig:main}.

\subsubsection{Training the bottom-level policy}
\label{sec:inverse-dynamics-module}
Given the state embedding~$\state_t$ and sub-goal embedding
$\hgoal_{t+k} = \toppolicy_{k}(\state_t, \cmd_t; \theta_t)$, we want to optimize the parameter  $ \theta_b$ for bottom-level policy.
Let $\hact_t = \policy (\state_t, \hgoal_{t+k}; \theta_b)$ be the action prediction
and $\act_t = \policy^*(\obs_t, \cmd_t)$ be the optimal action from the expert.

For our car experiments, the action space is composed of two actions $\ActSp = \ActSpS
\times \ActSpB$, where the steering angle is continuous $\hactst \in \ActSpS = [-1, 1]$, and the brake is binary $\hactbt \in \ActSpB = \{0,1\}$.
In this way, we have:
\begin{equation}
    L_{\text{Bottom}}(\hact_t, \act_t)=L_{\text{BCE}}(\hactbt,\actbt)+\lVert \actst - \hactst \rVert_2^2
    ,
\label{equ:l1}
\end{equation}
where $L_{\text{BCE}}$ is the Binary Cross Entropy loss.

\subsubsection{Training the top-level policy}
\label{sec:cond-trans-module}
As discussed earlier, we want to predict the future sub-goal from the current state using a top-level policy $\hgoal_{t+k} = \toppolicy_k (\state_t, \cmd_t;
\theta_t)$. Since we do not have access to the true sub-goal embedding, we
generate a sub-goal embedding from a future achieved observation $\obs_{t+k}$
and treat it as the true sub-goal embedding $\goal_{t+k} =
\goalencoder (\encoder(\obs_{t+k}; \theta_e); \theta_g)$.

It is important to note why we prefer Triplet Network over other options. We
critique two na\"ive approaches: Auto-encoder approach and Direct minimization
approach.

\noindent \textit{Auto-encoder} approach can be summarized as training
auto-encoders to predict goal-images.
However, the prediction in the image space is undesirable because it is unnecessarily hard and not useful for the purpose of describing task-goals. 
For example (from~\citep{pathak_curiosity-driven_2017}),
there is no use predicting tree leafs in a breeze, when instructing a car to go near the tree.

\noindent \textit{Direct minimization} approach minimizes
$\lVert \toppolicy (\state_t, \cmd_t; \theta_t)-\goalencoder (\state_{t+k}; \theta_g) \rVert$ directly.
Although this approach is intuitive, it does not give desirable results.
The reason is that there exists a trivial solution when the network outputs
a constant embedding irrespective of the input, which makes the loss to be always zero.

\begin{figure}[t]
  \centering
  \def\svgwidth{\linewidth}
\begingroup%
  \makeatletter%

  \providecommand\rotatebox[2]{#2}%
  \ifx\svgwidth\undefined%
    \setlength{\unitlength}{225.37143737bp}%
    \ifx\svgscale\undefined%
      \relax%
    \else%
      \setlength{\unitlength}{\unitlength * \real{\svgscale}}%
    \fi%
  \else%
    \setlength{\unitlength}{\svgwidth}%
  \fi%
  \global\let\svgwidth\undefined%
  \global\let\svgscale\undefined%
  \makeatother%
  \begin{picture}(1,0.23681542)%
    \put(0,0){\includegraphics[width=\unitlength,page=1]{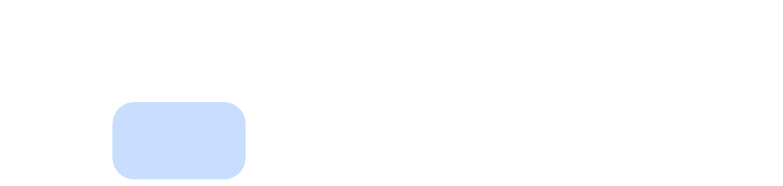}}%
    \put(0.00587832,0.02562375){\color[rgb]{0,0,0}\makebox(0,0)[lb]{\smash{$\obs_t$}}}%
    \put(0.00062716,0.19928534){\color[rgb]{0,0,0}\makebox(0,0)[lt]{\begin{minipage}{0.63894521\unitlength}\raggedright $\cmd_t$\end{minipage}}}%
    \put(0,0){\includegraphics[width=\unitlength,page=2]{triplet-network.pdf}}%
    \put(0.89808373,0.18986798){\color[rgb]{0,0,0}\makebox(0,0)[lt]{\begin{minipage}{0.40567947\unitlength}\raggedright $\obs_{t+k}$\end{minipage}}}%
    \put(0.65815742,0.20763449){\color[rgb]{0,0,0}\makebox(0,0)[lt]{\begin{minipage}{0.67570993\unitlength}\raggedright $\goalencoder(\encoder(.))$\\ \end{minipage}}}%
    \put(0,0){\includegraphics[width=\unitlength,page=3]{triplet-network.pdf}}%
    \put(0.52515636,0.22929303){\color[rgb]{0,0,0}\makebox(0,0)[lt]{\begin{minipage}{0.70993912\unitlength}\raggedright $\goal_{t+k}$\\ \end{minipage}}}%
    \put(0.3515747,0.05899875){\color[rgb]{0,0,0}\makebox(0,0)[lt]{\begin{minipage}{0.70993912\unitlength}\raggedright $\hgoal_{t+k}$\\ \end{minipage}}}%
    \put(0.20061545,0.08073864){\color[rgb]{0,0,0}\makebox(0,0)[lt]{\begin{minipage}{0.70993912\unitlength}\raggedright $\encoder$\end{minipage}}}%
    \put(0.25440147,0.2133387){\color[rgb]{0,0,0}\makebox(0,0)[lt]{\begin{minipage}{0.70993912\unitlength}\raggedright $\toppolicy_k$\end{minipage}}}%
    \put(0,0){\includegraphics[width=\unitlength,page=4]{triplet-network.pdf}}%
    \put(0.88151288,0.06173533){\color[rgb]{0,0,0}\makebox(0,0)[lt]{\begin{minipage}{0.40567947\unitlength}\raggedright $\obs_{r}$\end{minipage}}}%
    \put(0.65420873,0.07778102){\color[rgb]{0,0,0}\makebox(0,0)[lt]{\begin{minipage}{0.65669369\unitlength}\raggedright $\goalencoder(\encoder(.))$\\ \end{minipage}}}%
    \put(0,0){\includegraphics[width=\unitlength,page=5]{triplet-network.pdf}}%
    \put(0.52429793,0.05490621){\color[rgb]{0,0,0}\makebox(0,0)[lt]{\begin{minipage}{0.70993912\unitlength}\raggedright $\goal_{r}$\\ \end{minipage}}}%
    \put(0.46208333,0.19158757){\color[rgb]{0,0,0}\makebox(0,0)[lt]{\begin{minipage}{0.17748478\unitlength}\raggedright $d_+$\end{minipage}}}%
    \put(0.45954781,0.0759689){\color[rgb]{0,0,0}\makebox(0,0)[lt]{\begin{minipage}{0.21551726\unitlength}\raggedright $d_{-}$\end{minipage}}}%
    \put(0.45506565,0.1458907){\color[rgb]{0,0,0}\makebox(0,0)[lt]{\begin{minipage}{0.47540561\unitlength}\raggedright $L_{\text{Triplet}}$\end{minipage}}}%
    \put(0,0){\includegraphics[width=\unitlength,page=6]{triplet-network.pdf}}%
    \put(0.29695336,0.11081644){\color[rgb]{0,0,0}\makebox(0,0)[lb]{\smash{$\state_t$}}}%
  \end{picture}%
\endgroup%

  \caption{We train goal-embeddings by using Triplet Network~\cite{hoffer_deep_2014}.
  The matching pair of goal-embeddings $(\hgoal_{t+k}, \goal_{t+k})$ is considered as a positive example, while a random goal-embedding $\goal_r$ is used to construct a negative example $(\hgoal_{t+k}, \goal_{r})$. Refer to Eq~\eqref{equ:triplet} for details.}
  \label{fig:triplet-net}
\end{figure}

Our goal is to assign high similarity to the pair of states that is temporally aligned, and differentiate all other $\obs_\tau :\tau \ne t+k$ in the goal embedding space. 
This is the main idea behind Triplet Loss~\citep{hoffer_deep_2014}.
In Figure~\ref{fig:triplet-net}, each training sample contains two triples $[(\state_t,\state_{t+k},\lclose_1),(\state_t,\state_r,\lclose_2)]$, and each triple consists of two observations and a binary label.
Two observations are considered close $\lclose_1 = 1$ if they are separated by exactly $k$ time steps.
In order to differentiate these two states with all others, we add a random observations $\state_r$ with label $y_2 = 0$.
This structure is shown in Fig.~\ref{fig:triplet-net}, and we have:
\begin{align}
  \text{TripletNet}(\state_t,\state_{t+k},\state_r)
  &=
\begin{bmatrix} 
\underbrace{\lVert \toppolicy_{k}(\state_t, \cmd_t)-\goalencoder (\state_{t+k}) \rVert}_{d_+} \\
\underbrace{\lVert \toppolicy_{k}(\state_t, \cmd_t)-\goalencoder (\state_{r}) \rVert}_{d_{-}}
\end{bmatrix},
  \\
  L_\text{Triplet}(\state_t, \state_{t+k}, \state_r) &= \left (
                                          \frac{\exp (d_+)}
                                          {\exp (d_+)
                                          +\exp (d_{-})}
                                          \right)^2.
\label{equ:triplet}
\end{align}
The idea of Triple-Net is illustrated in Figure~\ref{fig:triplet-net}.
Each of the input observation $\obs_t$ is embedded as a vector $\encoder (\obs_t)$ using an encoder based on
 ResNet50~\citep{res}.

By learning the bottom-level policy and top-level policy jointly, we yield to the following objective function:
\begin{align}
\label{equ:loss}
\theta^*_t, \theta^*_g, \theta^*_b= &\arg\min_{\theta_{t,g,b}}
\sum_{t \in D}
\{L_{\text{Bottom}}(\policy (\state_t, \hgoal_{t+k, \theta_t}; \theta_b), \act_t) \nonumber\\
&+ L_{\text{Bottom}}(\policy (\state_t, \goal_{t+k,\theta_g}; \theta_b), \act_t) \nonumber\\
&+ L_{\text{Triplet}}(\state_t, \state_{t+k}, \state_r; \theta_t,  \theta_g)\} 
.
\end{align}%

\subsubsection{Desired policy approximation}
The main claim of our work is that using hierarchical sub-goals helps in constructing a better desired policy $\policy_{\text{des}}$.
As constructed above, we assume that we can factorize the
$\policy_{\text{des}}$ into two levels:
\begin{align}
\begin{cases}
  \policy_{\text{des}}(\state_t, \cmd_t)&\defeq \policy_{\text{eff}}(\state_t, \goal_{t+k,\theta_g}),
  \\
  \goal_{t+k,\theta_g} & = \goalencoder(\state_{t+k}; \theta_g).
 \end{cases}
\end{align}%
In order to achieve a good approximation for above functions, we need two assumptions. 
Firstly, we assume that reaction-delay of the expert does not change the achieved state
$\state_{t+k}$ down the trajectory (as in Fig.~\ref{fig:paper-summary}).
Secondly, we need to assume that the lower-level policy $\policy_{\text{eff}}(\state_t, \goal_{t+k,\theta_g})$ following
sub-goals can be learned more easily and in fewer iterations than the full policy. 
These two assumptions depend on the value of $k$. 
The first assumption holds well for large $k$s.
For example, a large $k$ means that we choose a far away $\state_{t+k}$ as a proxy for the goal.
Even with a large expert reaction delay, the trajectory is likely to return to desired trajectory by time $t+k$.
The second assumption holds well for small $k$s. Consider the example of $k=1$, where the direction of movement can almost be inferred by optical flow. 
Because of this trade-off on $k$, we conduct experiments to evaluate the effect of $k$ on our algorithm. 
The details are discussed in the Experiments section.

\subsection{Learning from intervention by Backtracking}
To construct a better approximation of the expert's desired behavior, we collect data not only at the point of intervention but also for a few time steps around it.
Therefore, during the data collection phase, we keep track of past $M$ time-steps in a data queue, $D_\beta$.
Whenever the intervention happens at time $t$, we interpolate the actions between $\act_{t-M}$ and the intervened action $\act^*_t$ to update the action in data-queue $D_\beta$. 
The Backtracking function, $\text{Backtrack}(\act^*_{t}, \act_{t-M}, j)$, interpolates the actions to time step $t-j$.
This data-collection algorithm is summarized in Alg~\ref{alg:hlid}.

\begin{algorithm}[h]
\hspace*{0.02in} {\bf Input:} 
Parameter $M$: the number of time-steps before intervention time $t$ s.t. $\act_{t-M}$ is a desired action, 
maximum episode length $T$, policies $\policy_k(\state_t, \cmd_t)$, and the demonstration data from expert $D_{h}$ \\
\hspace*{0.02in} {\bf Output:} Updated policies  $\policy_{k+1}(\state_t, \cmd_t) $
\begin{algorithmic}[1]
\caption{Learn-form-Intervention by Backtracking}
\label{alg:hlid}
\Repeat
  \State Sample a new state $\state_0$ from the environment
  \State Initialize data queue $D_\beta \gets \emptyset$ $D_\alpha \gets \emptyset$
  \For{$ t= 1,\dots,T$}
    \State Sample a new state $\state_t$ from the environment\;
      \If {no Intervention}
        \State Execute $\act_t=\policy_k(\state_t, \cmd_t)$
        \State Append $(\state_t , \act_t)$ to $D_\beta$
        \If{ len$(D_\beta) > M$} 
            \State POP$(D_\beta)$ 
        \EndIf 
        \If{len$(D_\alpha) > 0$} 
         \State Append $D_h \gets D_h \bigcup \{D_\alpha\}$
         \State  $\alpha \gets \emptyset$
        \EndIf 
      \Else{}
        \State Execute $\act^*_t=\policy^*(\state_t, \cmd_t)$
        \For{$j = 1, \dots, M$}
            \State $\state_{t-M}, \act_{t-M} \gets D_\beta[M]$
            \State $D_\beta[j][\act] \gets \text{Backtrack}(\act^*_t, \act_{t-M}, j$)
        \EndFor
        \State Append $D_h \gets D_h \bigcup \{D_\beta\}$
        \State Append $(\state_t , \act_t)$ to $D_\alpha$
        \State  $D_\beta \gets \emptyset$
        \EndIf
  \EndFor
\Until End of collecting 
\State Update policies $\policy_{k+1} \gets $Train$(\policy_{k},D_{h})$
\end{algorithmic}
\end{algorithm}
\section{Experimental Setup}
\subsection{Simulated Environment}   
We use a 3D urban driving simulator CARLA~\citep{CARLA}.
It offers a realistic emulation of vehicle dynamics and information, including distance traveled, collisions, and the occurrences of infractions (e.g., drifting into the opposite lane or the sidewalk).
By employing the agent in CARLA simulator, we are able to automate the training and evaluation process.

\subsection{Data Collection}
The agent is initialized at a random location with 80 other vehicles in town at each iteration.
A local-planner was used to generate roaming routes containing waypoints using topological commands, $\cmd_t$ (e.g., \texttt{TURN\_LEFT}), provided by a human expert. 

The expert policy is presented by a mixture of a Proportional-Integral-Derivative (PID) controller and a human. PID controller uses the route provided by the planner to compute the steer-angle. A human expert is presented with a third-person view of the environment, who takes control over the PID controller when the simulated vehicle stuck into unrecoverable states. The expert intervenes only in either of the following situations, potential crash or potential lane evasion. For now, we ignore traffic lights and stop signs in order to keep the same setting with \cite{codevilla_end--end_2018}, but these can be easily added to our framework when needed.

As discussed earlier, the recorded control signal contains the steering angle $\actst$ and the brake $\actbt$. The steering angle is between $\ActSpS = [-1,1]$, with extreme values corresponding to full left and full right, respectively. The brake is a binary signal $\ActSpB = \{0, 1\}$, where 1 indicates full brake. The speed of the car is controlled by a separate PID controller at 50km/h. We don't record the acceleration signal because it depends on the instantaneous velocity of the car, which is subject to various factors and hard to model.

We also recorded topological commands $\cmd_t$ that are provided by either human or planner. 
We split all topological commands into three categories, $\Cmd = \{0, 1, 2\}$: \texttt{lane-following} ($\cmd_t = 0$), turning \texttt{left} ($\cmd_t = 1$) or \texttt{right} ($\cmd_t = 2$) at the intersection.

\subsection{Evaluation}
\subsubsection{Implementation details} 
The observation from CARLA at each time-step is an $800\times600$ resolution image.
For the baseline \textbf{Branched} model, the observation is first encoded into a 1024-dimension feature vector $\encoder(\state_t)$ using ResNet-50~\cite{res}. It consists of 5 stages of identity and convolution blocks, both of which has 3 convolution layers, respectively.

Three separate two-headed Multi-Layer Perceptrons (MLP) follow the Resnet-50 encoder. Each of them has 3 fully-connected (FC) layers followed by Exponential Linear Unit (ELU) non-linearity. 
This is used as the bottom-level policy $\policy(\state_t, \goal_{t+k})$.

Our top-level policy, $\toppolicy(\state_t, \cmd_t)$, again contains three separate MLPs that use command $\cmd_t$ as switch.
All other modules are implemented as standard MLPs.
We use ELU non-linearities after all hidden layers and applied 50\% dropout after fully-connected hidden layers
We initiate ResNet-50 with pre-trained parameters and only fine-tune the top three stages.
The Equation~\ref{equ:loss} is minimized with a learning rate of 1e-5 using Adam solver.

For each experiment, we use behavior cloning with 30 mins recorded data ($\sim$7200 frames) in our first iteration and test agent for 15 mins in each subsequent iteration. 
We vary two properties of the algorithms for evaluation: data collection and policy representation. We have three variations for the data-collection approach: 
\begin{itemize}
\item \textbf{Demo} stands for Behavior Cloning~\citep{survey} using intervention data as well as non-intervention data as additional demonstration data.
\item \textbf{CoL} stands for Cycle-of-Learning~\citep{goecks_efficiently_nodate}, which uses only intervention data as additional demonstration data and ignores the non-intervention data.
\item \textbf{LbB} stands for our approach, Learning from intervention by Backtracking, as described earlier in the Methods section. We ignore most of the non-intervention data, except $M$ steps before and after the intervention.
\end{itemize}
For policy representation, we use two variations:
\begin{itemize}
    \item 
    \textbf{Branched} denotes the best setting in \citep{codevilla_end--end_2018}, which uses a feature extractor followed by three parallel Multi-Layer Perceptrons (MLPs). The condition of the expert's intention acts as a switch that selects which MLP is used. This model does not use hierarchy.
    \item 
    \textbf{Sub-goal}  denotes our proposed hierarchical network structure with $k=5$ as described in the Methods section.
\end{itemize}

\section{Experiments and Results}
We perform three sets of experiments. The first two experiments evaluate the accuracy and data-usage of different algorithms over training iterations. The effect of the hyper-parameter $k$ (see Eq~\eqref{eq:hierarchical-policy}) on our proposed algorithm is evaluated in the third experiment.

\begin{figure}[t]
  \centering
  \includegraphics[width=\linewidth]{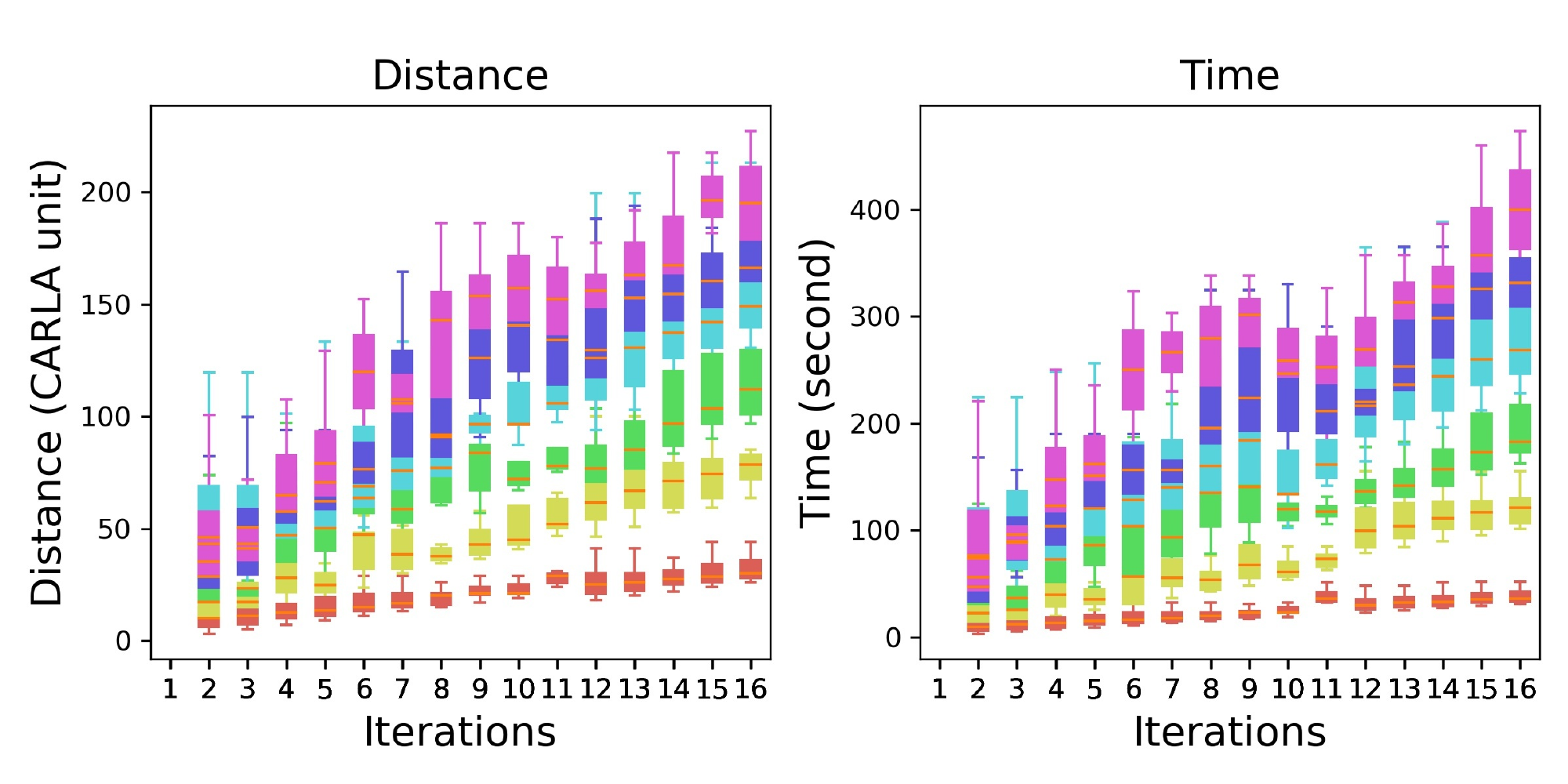} \\
  \includegraphics[width=\linewidth]{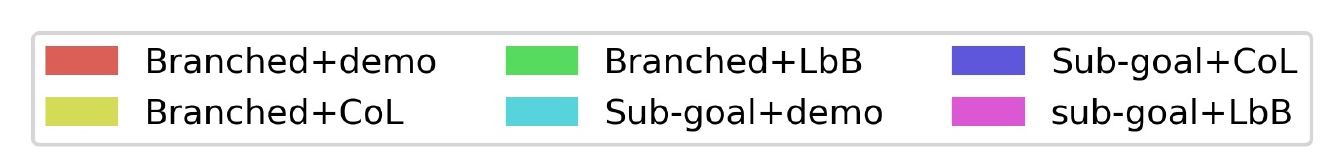}
  \caption{Graph showing the progress of the agent as it grows to be able to travel for longer distances and times, respectively, before needing expert intervention. For both graphs, higher is better. We compare the effects of our contributions (\textbf{Sub-goal} and \textbf{LbB}) with baseline approaches. The combination of our two contributions  \textbf{Sub-goal+LbB} produces the best results. }
  \label{fig:exp-accuracy}
\end{figure}

\begin{figure}[t]
  \centering
  \includegraphics[width=0.9\linewidth]{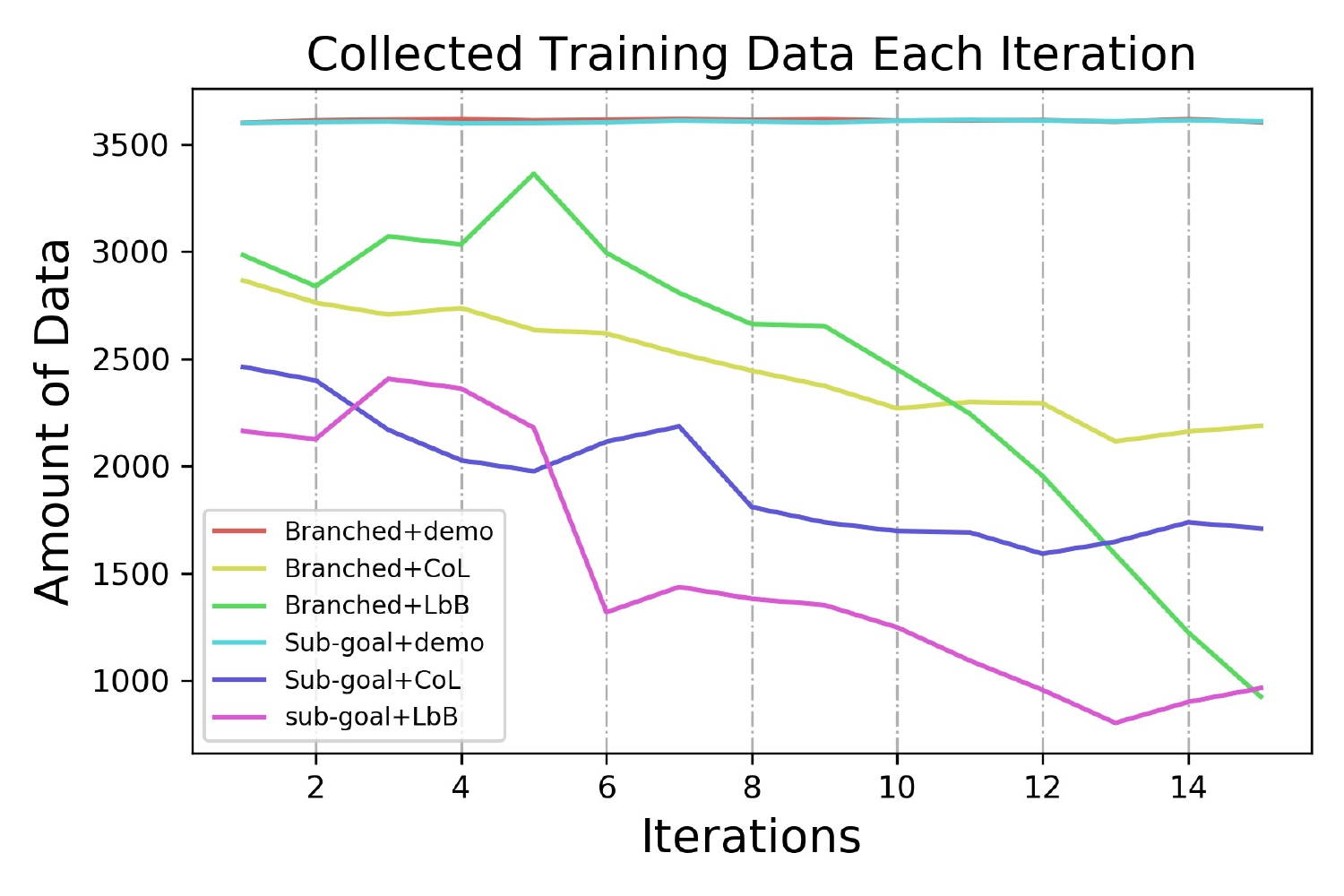} \\
  \includegraphics[width=0.9\linewidth]{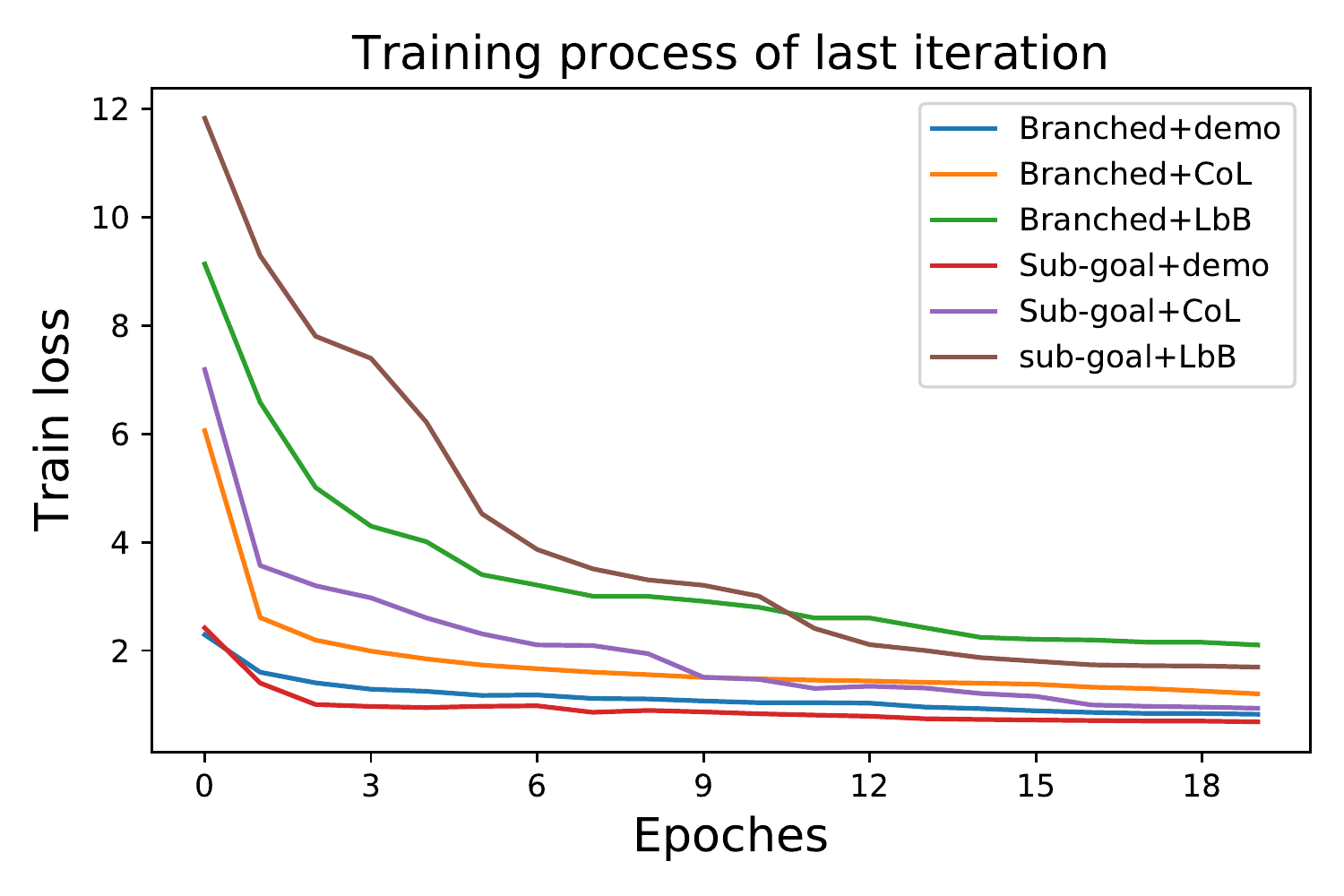}%
  \caption{\textbf{Top:} Comparison of the number of data-samples per iteration needed to train the various algorithms. The number of data-samples used is proportional to the number of expert interventions needed per iteration. Lower is better, indicating that the algorithm is learning from less training data. Our algorithm \textbf{Sub-goal+LbB} shows a sharp drop in the amount of data needed per iteration, followed by consistently low values.
  \textbf{Bottom:} Loss curves during training for different algorithms. Each algorithm uses different amount of data (proportional to number of interventions) during training. For fairer comparison refer to Figure~\ref{fig:exp-accuracy}. The plot is included for completeness.}
  \label{fig:exp-data-efficiency}
  \label{fig:training-loss}%
\end{figure}

\paragraph{Accuracy}
We evaluated our method in terms of the two metrics: the \textbf{time} and \textbf{distance} without intervention.
Time and distance without intervention measure is a standard measure of progress of autonomous driving.
In the absence of an expert overseer, it can be assumed to be the same as the duration of the agent successfully driving itself in autonomous mode. The results across different algorithms are shown in Figure~\ref{fig:exp-accuracy}.
Note that Behaviour Cloning, \textbf{Branched+demo}, has a moderate performance increase in the first few iterations due to the increase of dataset, but fails to keep increasing as the dataset increments.
The results also show a significant increase in performance due to both of our contributions \textbf{LbB} and \textbf{Sub-goal}.
\paragraph{Data efficiency}
In this experiment, we explore how much data each algorithm consumes as a function of intervention iterations. In Figure~\ref{fig:exp-data-efficiency}, we show that as the learning policy improves, the algorithm needs fewer interventions, hence less data.
 As \textbf{demo} uses all the data, including the data with interventions and without interventions, the number of data-samples is highest and stays constant per iteration.
 \textbf{CoL} and \textbf{LbB} use only intervention data, so the data-samples used depend upon the accuracy of the algorithm. Again due to faster-learning, our proposed method \textbf{sub-goal+LbB} uses the least amount of data.
 Our results also confirm results from~\cite{goecks_efficiently_nodate} that Learning from Intervention (LfI) is data-efficient, as it uses only the intervention data rather than all data.
As the demonstrations continue, it becomes more likely to encounter seen sates and the states where the algorithm already performs well.
LfI allows the algorithm to focus on critical intervention-related states and learn corrective actions.

To confirm the idea that the number of data samples is proportional to the number interventions, we calculate the mean and variance of the number of frames per intervention for six algorithms:\textbf{branched+demo}, \textbf{branched+CoL}, \textbf{branched+LbB}, \textbf{sub-goal+demo}, \textbf{sub-goal+CoL}, \textbf{sub-goal+LbB}.
We find the mean = $[22, 17, 23, 20, 23, 25]$ (4 frame/second)  and the variance = $[4, 6, 5, 6, 3, 4]$ in that order. 
The low variance shows that the total number of intervention is roughly proportional to the data added, and will have a similar decay as shown in Figure~\ref{fig:exp-data-efficiency} for the number of interventions.
\paragraph{Training loss}
We show the training curves for a single intervention batch in Figure~\ref{fig:training-loss}. 
We make two observations. First, \textbf{Sub-goal} uses less data than \textbf{Branched}, which leads to a slower convergence but eventually leads to better accuracy. 
Secondly, \textbf{LbB} results in bigger initial loss than other data-collection methods (\textbf{CoL} or \textbf{Demo}). 
We think this is because \textbf{LbB} is able to collect more relevant data than \textbf{Demo} but 
yet produces more data-samples than \textbf{CoL}. 

Please note that these loss curves do not reflect the performance of the algorithms.
This is because of two reasons. First, these loss curves are for the last iteration of the many training cycles performed during full online training.
Second, different models have different data sizes (based on the number of interventions) after the initial iteration, so the comparison might seem misleading. 

\begin{figure}[t]
  \centering
  \includegraphics[width=\linewidth]{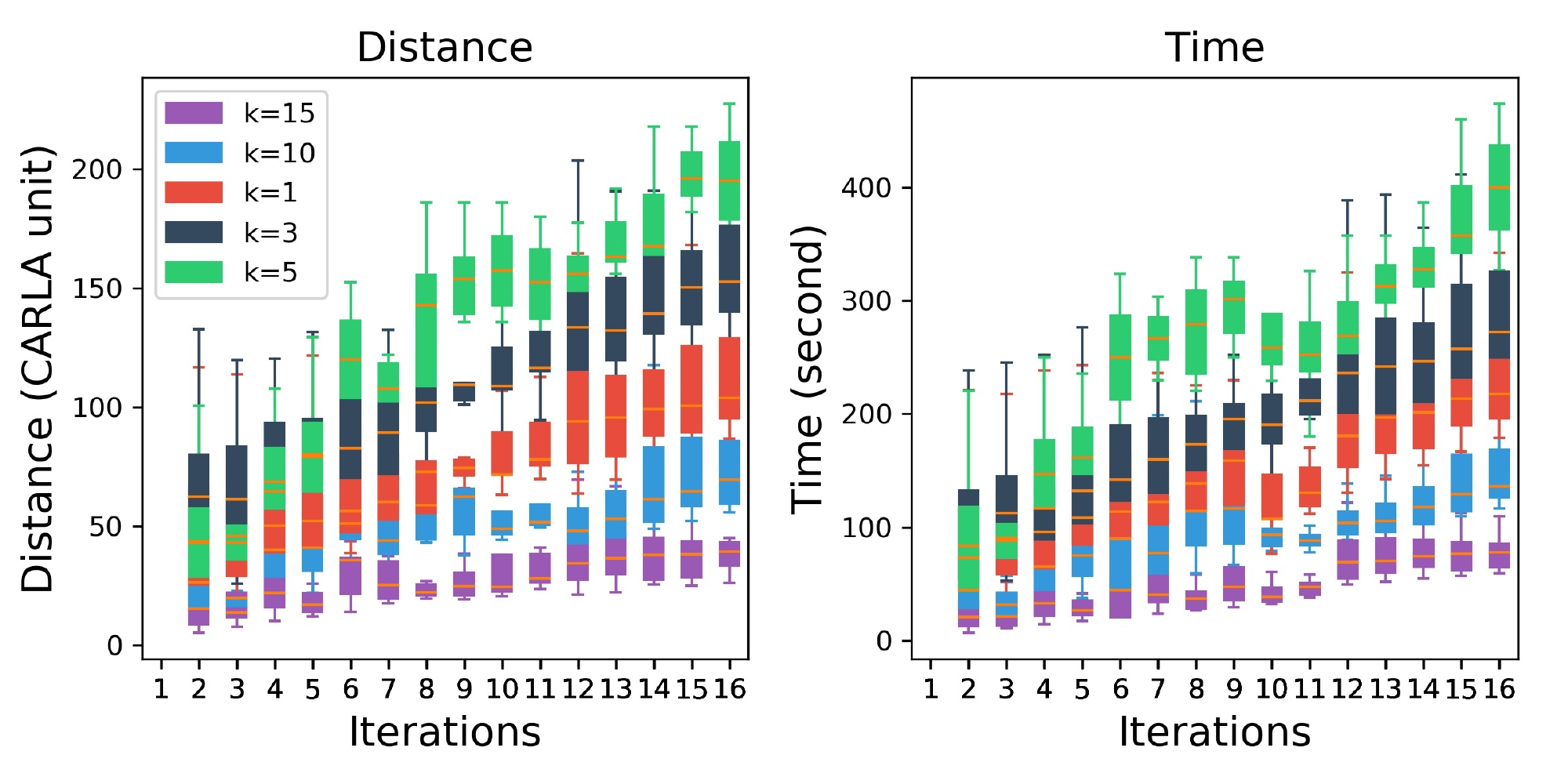}
  \caption{Evaluation of the effect of $k$ on our proposed \textbf{Sub-goal+LbB} algorithm. The graph shows the time and distance traveled by the agent without intervention. Higher is better. Matching our expectations, the results show that $k$ represents a trade-off between predicting the sub-goals far into the future (high $k$) by the top-level policy with the ability of the bottom-level policy to follow sub-goals correctly (low $k$). $k=5$ provided the best performance, so we continued to use that value for the rest of our experiments. }
  \label{fig:exp-hyper-param-k}
\end{figure}

\paragraph{Effect of the hyper-parameter $k$}
We also investigate the effect of the hyper-parameter $k$ in
Eq~\eqref{eq:hierarchical-policy}, which represents the trade-off between
the long-term sub-goal prediction and robustness to the expert's reaction time. Our
dataset is captured at 4 frames per second, hence $k=4$ is equivalent to 1
second. The results are shown in Figure~\ref{fig:exp-hyper-param-k}.

We evaluate the effect of the hyper-parameter $k$ on our proposed algorithm \textbf{Sub-goal+LbB}.
As we discussed above, the parameter $k$ indicates how far in the future the top-level policy is learning to predict the sub-goal $\goal_{t+k}$.
We choose $k \in \{1,3,5,10,15\}$ which is 0.25s to 3.75 second in the future.
The results match our intuition that $k$ represents a trade-off between predicting sub-goal far into the future 
by top-level policy while the ability of bottom-level policy to follow it correctly.
With a small value of $k$, like $k = 1$, the policy learned is reactive and similar to the baseline \texttt{Branched}.
Increasing $k$ up to a point improves the performance, but after a point, the bottom-level policy is unable to learn to follow the far-off goals correctly. We find $k=5$ as the sweet spot and use it for all other experiments.
%
\section{Real-World Demonstration}

\begin{figure}[t]
  \centering
  \includegraphics[width=0.8\linewidth]{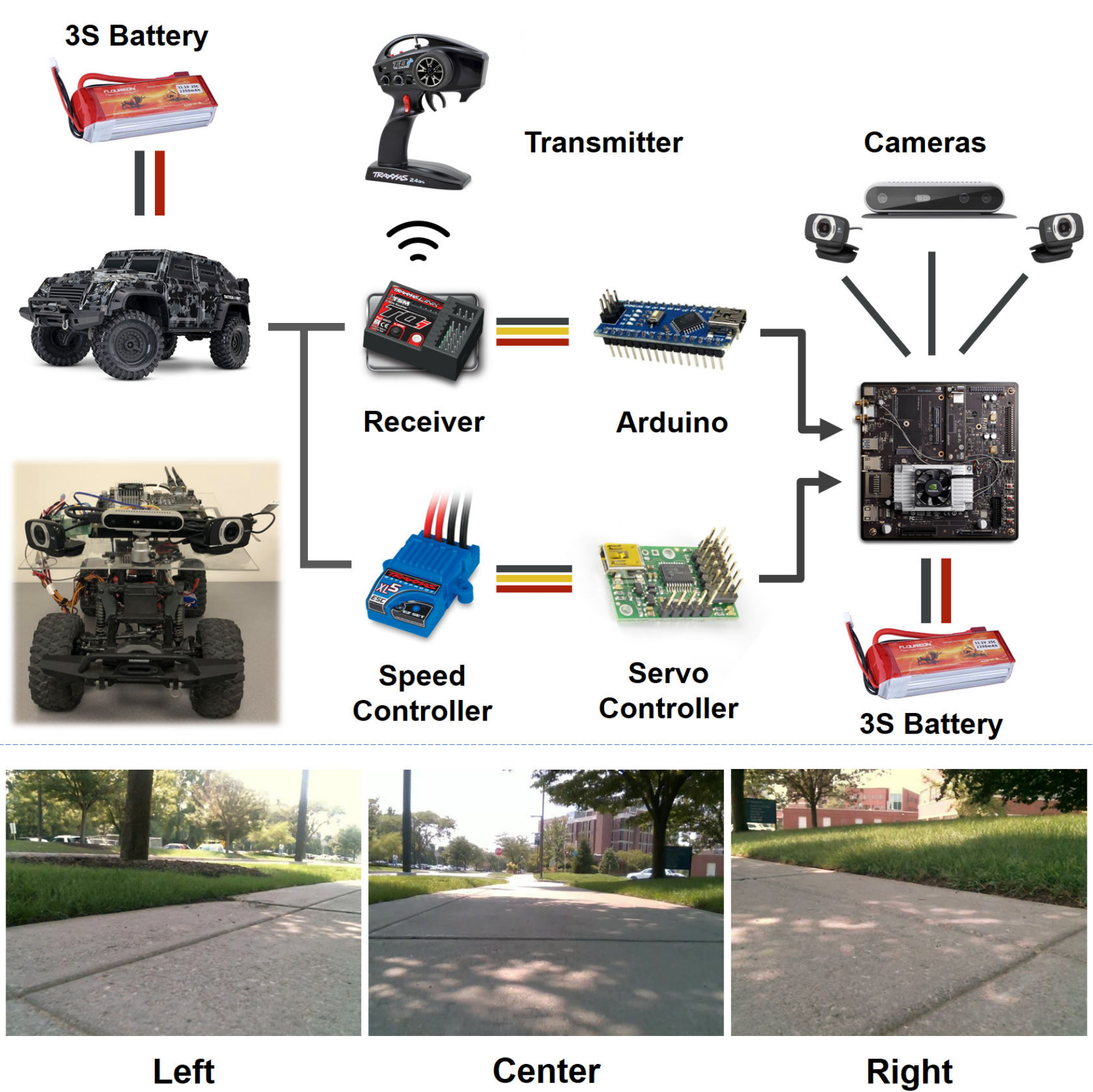}
  \caption{The architecture of our hardware system setup indicating how wires connect and frames captured by three cameras in one spot. This three-camera setting provides a wide field of view to ensure the sides of paths can be recorded.}
  \label{fig:hardware}
\end{figure}

To demonstrate the practicality of our approach, we deploy \textbf{Branched+LbB} on a real RC car with a few differences. 
Instead of a car rich environment, we study navigation in a pedestrian-rich environment. 
Also, we have different command categories (or scenarios):~\{\texttt{path-following} (with no pedestrian),~\texttt{pedestrian-following},~\texttt{confronting} (avoid hitting a confronting person), ~\texttt{crossing} (avoid hitting a crossing pedestrian)\}.
We equipped an off-the-shelf 1/10 scale ($13''\times10''\times 11''$) truck with an embedded computer (Nvidia TX2), an Intel RealSense D415 as the primary central camera and two webcams on the sides. 
The setup of the physical system is shown in Figure~\ref{fig:hardware}. 
We also employ an Arduino bboard and a USB servo controller as our control system to the robot.
The decrease in the amount of data required, corresponding to the number of
interventions needed, is shown in Figure~\ref{pic:data}. We include the demo
video in the supplementary material.
\begin{figure}[t]
\centering
	\includegraphics[width=0.8\linewidth]{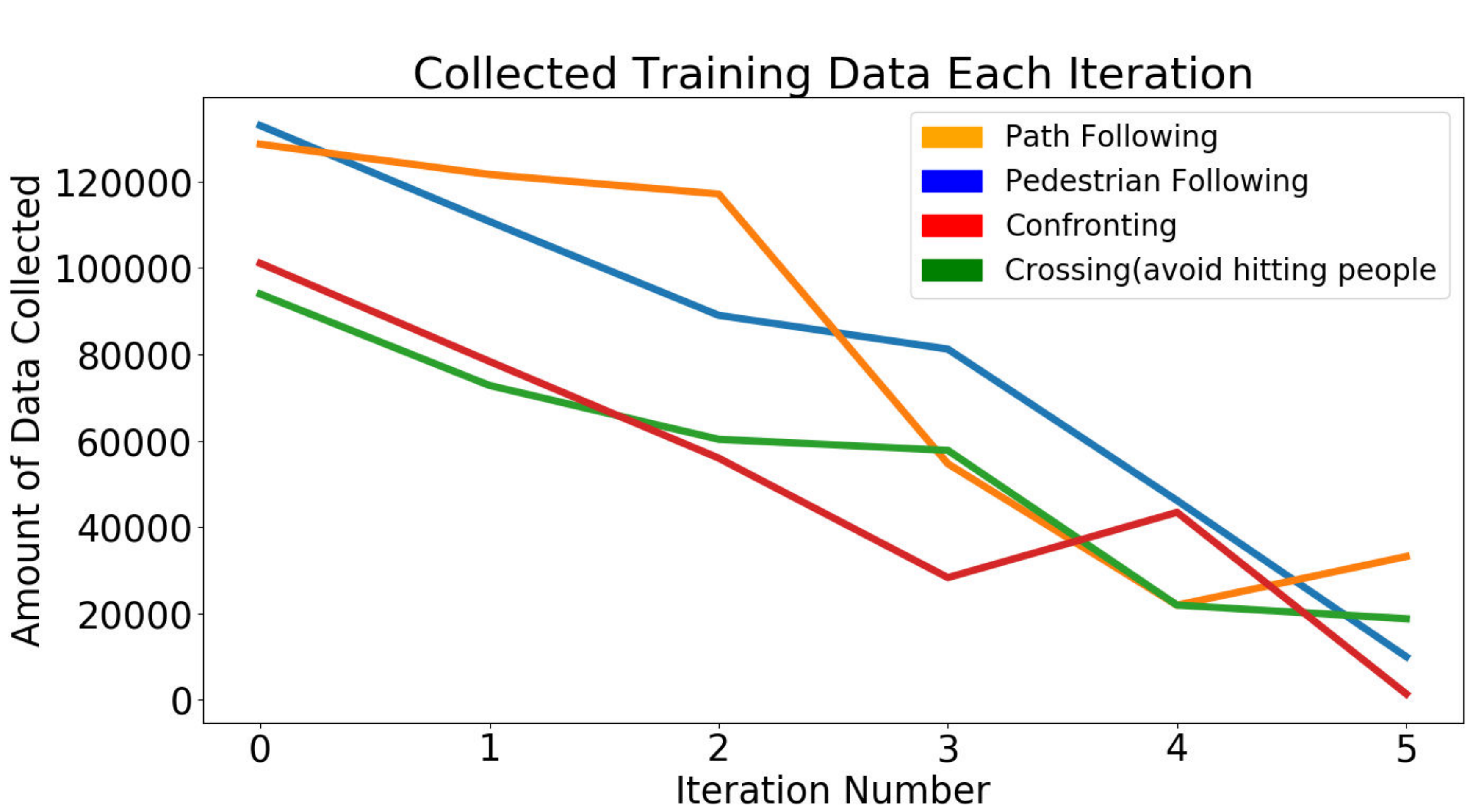}
	\caption{The amount of data collected (number of frames) for the real RC car experiment. Changes in size of data-samples per iteration, proportional to the number of interventions, used in four scenarios. Lower is better.}
	\label{pic:data}
\end{figure}

\section{Conclusions and Future Work}
To conclude, we introduced a new problem formulation that accounts for the expert's reaction delay in Learning From Interventions (LfI).
We proposed a new method to solve the proposed variation of LfI, which combines LfI with Hierarchical RL.
We implemented the method on an autonomous driving problem in CARLA simulator using a Triplet-Network based architecture. 
We also proposed an interpolation trick called Backtracking, that allows us to use state-action pairs before and after the intervention.
Our experiments show the value of both innovations.
Our experiments also confirm the superiority of LfI approaches over Learning from Demonstrations in terms of data efficiency.
Additionally, experimenting with the values of $k$ provides insight into how we can help the agent understand the environment, by allowing it to consider the temporal relation of subsequent states.

While the presented results are encouraging, there also remains significant room for progress.
Learning from Interventions can easily be extended to IRL, which will allow us to improve data-efficiency further while ensuring safety. 
Learning from Intervention is timely, because there already exist Level-3 autonomy cars on the roads, with abundantly available intervention data that can be used with our approach to training the algorithms further. Finally, our algorithm has the limitation of resorting to a heuristic-based approximation of the effective policy. In future work, we will devise a principled way of coming up with that approximation.

\section{Acknowledgments}
J. Bi, T. Xiao, and C. Xu are supported by NSF IIS 1741472, IIS 1813709, and NIST 60NANB17D191 (Sub). V. Dhiman is supported by the Army Research Laboratory - Distributed and Collaborative Intelligent Systems and Technology Collaborative Research Alliance (DCIST-CRA). This article solely reflects the opinions and conclusions of its authors but not the funding agents.

\bibliographystyle{aaai}

\end{document}